\documentclass[10pt,twocolumn,letterpaper]{article}

\usepackage{iccv}
\usepackage{times}
\usepackage{epsfig}
\usepackage{graphicx}
\usepackage{amsmath}
\usepackage{amssymb}

\usepackage{booktabs}
\usepackage{multirow}
\usepackage{rotating}
\usepackage{subcaption}

\graphicspath{{figures/}}

\newcommand{\Tref}[1]{Table~\ref{#1}}
\newcommand{\Eref}[1]{Eq.~(\ref{#1})}
\newcommand{\Fref}[1]{Fig.~\ref{#1}}

\newcommand{\Sref}[1]{Section~\ref{#1}}

\usepackage[breaklinks=true,bookmarks=false]{hyperref}

\iccvfinalcopy 

\def\iccvPaperID{5455} 

\begin{document}

\title{Unsupervised Out-of-Distribution Detection by Maximum Classifier Discrepancy}

\author{Qing Yu\ \ \ \ Kiyoharu Aizawa\\
The University of Tokyo, Japan\\
{\tt\small \{yu, aizawa\}@hal.t.u-tokyo.ac.jp}
}

\maketitle


\begin{abstract}
   Since deep learning models have been implemented in many commercial applications, it is important to detect out-of-distribution (OOD) inputs correctly to maintain the performance of the models, ensure the quality of the collected data, and prevent the applications from being used for other-than-intended purposes. In this work, we propose a two-head deep convolutional neural network (CNN) and maximize the discrepancy between the two classifiers to detect OOD inputs. We train a two-head CNN consisting of one common feature extractor and two classifiers which have different decision boundaries but can classify in-distribution (ID) samples correctly. Unlike previous methods, we also utilize unlabeled data for unsupervised training and we use these unlabeled data to maximize the discrepancy between the decision boundaries of two classifiers to push OOD samples outside the manifold of the in-distribution (ID) samples, which enables us to detect OOD samples that are far from the support of the ID samples. Overall, our approach significantly outperforms other state-of-the-art methods on several OOD detection benchmarks and two cases of real-world simulation.

\end{abstract}

\section{Introduction}
    After several breakthroughs of deep learning methods, deep neural networks (DNNs) have achieved impressive results and even outperformed humans in fields such as image classification \cite{he2016deep}, face recognition \cite{Liu_2017_CVPR}, and natural language processing \cite{devlin2018bert}. Meanwhile, increasingly more commercial applications have implemented DNNs in their systems for solving different tasks with high accuracy to improve the performance of their products.

    To achieve stable recognition performance, the inputs of these models should be drawn from the same distribution as the training data that was used to train the model \cite{zhang2016understanding}. However, in the real-world, the inputs are uploaded by users, and thus the application can be used in unusual environments or be utilized for other-than-intended purposes, which means that these input samples can be drawn from different distributions and lead DNNs to provide wrong predictions. Therefore, for these applications, it is important to accurately detect out-of-distribution (OOD) samples.

    \begin{figure}
        \centering
        \includegraphics[width=8.4cm]{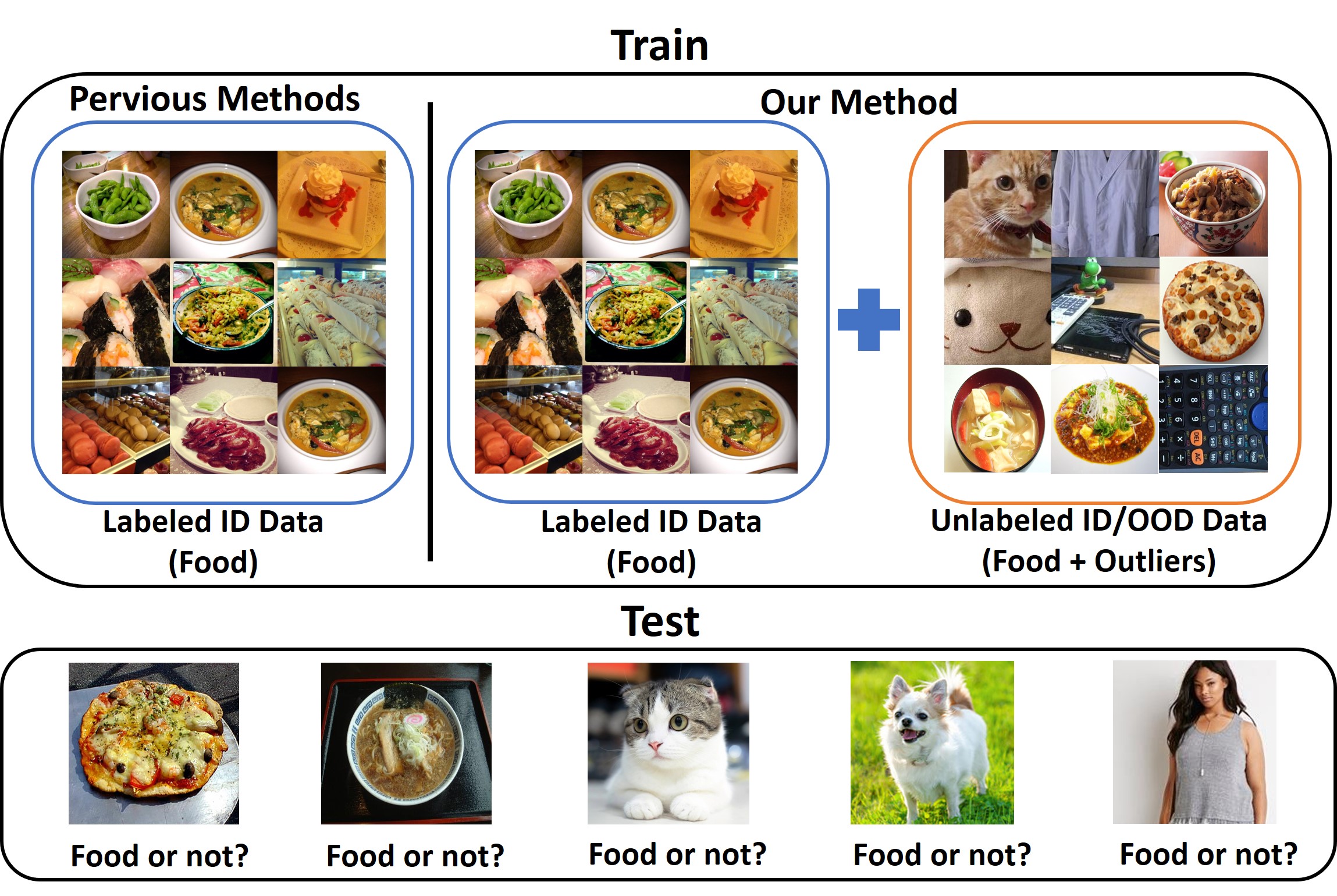}
        \vspace{-20pt}
        \caption{Experimental settings of OOD detection. Our method utilizes both labeled ID data and unlabeled ID/OOD data for training, which is different from previous methods. Please notice that we do not know which semantic class the unlabeled sample belongs to or whether the unlabeled sample is ID or OOD.}
        \label{fig:data}
        \vspace{-17pt}
    \end{figure}

    In this work, we propose a new setting for unsupervised out-of-distribution detection. While previous studies \cite{hendrycks2016baseline, lee2017training, liang2017enhancing, shalev2018out, vyas132018out} only use labeled ID data to train the neural network under supervision, we also utilize unlabeled data in the training process. \Fref{fig:data} shows our experimental settings of OOD detection for food recognition. Though we do not know the semantic class of the unlabeled sample or whether the unlabeled sample is ID or OOD, we find that these data are helpful for improving the performance of OOD detection and this kind of unlabeled data can be obtained easily in real-world applications.
     	
\begin{figure}
        \centering
        \includegraphics[width=8.4cm]{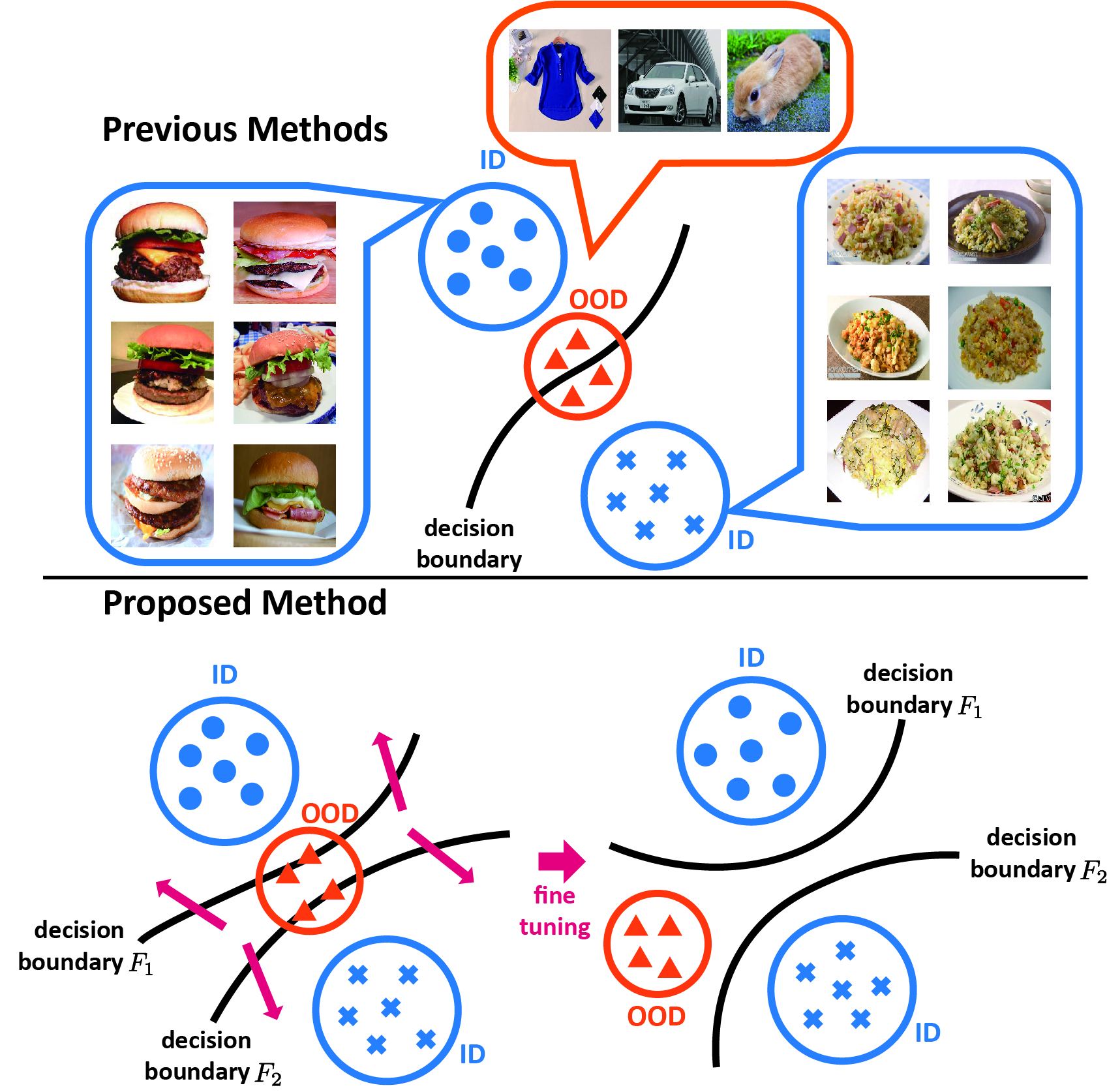}
        \vspace{-15pt}
        \caption{Comparison of previous and the proposed OOD detection methods. \textbf{Upper}: Previous methods try to detect OOD samples via the distance from the decision boundary. \textbf{Lower}: The proposed method detects OOD samples by the discrepancy between two classifiers.}
        \label{fig:overview}
        \vspace{-20pt}
    \end{figure}

   To utilize these unlabeled data, we also propose a novel out-of-distribution detection method for DNNs. Many OOD detection algorithms attempt to detect OOD samples using the confidence of the classifier \cite{hendrycks2016baseline, lee2017training, liang2017enhancing, vyas132018out}. For each input, a confidence score is evaluated based on a pre-trained classifier, then the score is compared to a threshold, and a label is assigned to the input according to whether the confidence score is greater than the threshold. Those samples having lower confidence scores (which means they are closer to the decision boundary) are classified as OOD samples, as shown in the upper part of \Fref{fig:overview}.  
    In the previous works, they used CIFAR-10/CIFAR-100 as ID and other datasets, TinyImageNet/LSUN/iSUN as OOD. Though there is a small overlap of classes between ID and OOD, we follow the same setting for comparisons.

    Although the existing methods are effective on some datasets, they still exhibit poor performance when ID dataset has big class numbers. For example, using CIFAR-100 \cite{Krizhevsky09learningmultiple} (a natural image dataset with 100 classes) as the in-distribution (ID) dataset and TinyImageNet \cite{imagenet_cvpr09} (another natural image dataset with 200 classes) as the OOD dataset, which means current methods cannot separate the confidence scores of ID samples and OOD samples well enough.

    To overcome this problem, we introduce a two-head deep convolutional neural network (CNN) with one common feature extractor and two separated classifiers as shown in \Fref{fig:method}. Since OOD samples are not clearly categorized into classes of ID samples or far from the distribution of ID samples, the two classifiers having different parameters will be confused and output different results. Consequently, as shown in the lower left part of \Fref{fig:overview}, OOD samples will exist in the gap of the two decision boundaries, which makes it easier to detect OOD samples. To achieve better performance, we further fine-tune the neural network to correctly classify labeled ID samples, and maximize the discrepancy between the decision boundaries of two classifiers, simultaneously. Please note that we do not use labeled OOD samples for training in our method.

    We evaluate our method on a diverse set of in- and out-of-distribution dataset pairs. In many settings, our method outperforms other methods by a large margin. The contributions of our paper are summarized as follows:
    \begin{itemize}
        \vspace{-5pt}
        \setlength\itemsep{0pt}
        \item We propose a novel experimental setting and a novel training methodology for out-of-distribution detection in neural networks. Our method does not require labeled OOD samples for training and can be easily implemented on any modern neural architecture.

	    \item We propose utilizing the discrepancy between two classifiers to separate in-distribution samples and OOD samples.
        
        \item We evaluate our method on state-of-the-art network architectures, such as DenseNet \cite{huang2017densely} and Wide ResNet (WRN) \cite{Zagoruyko2016WRN}, across several out-of-distribution detection tasks, including not only several OOD detection benchmarks but also real-world simulation datasets.
      \end{itemize}

\section{Related Work}
\begin{table*}[t]
        \centering
        \caption{Summary of recent related methods. Model change is how the method modifies the original classification network. Test complexity equals the required number of passing over the network times the number of the network. Training data is the type of data used for training by each method. AUROC is the area under receiver characteristic curve (detailed in \Sref{sec:exp}). Performance is shown for DenseNet trained on CIFAR-100 as ID and TinyImageNet-resized as OOD.} 
        \vspace{-10pt}
		\label{tbl:comp}
        \scalebox{0.9}{
            \begin{tabular}{c|c|c|c|c|c}
                \toprule
                 Method & Input pre-processing & Model change & Test complexity & Training data  & AUROC \\
                 \midrule
                 Hendrycks \& Gimpel \cite{hendrycks2016baseline} & No & No & 1 & Labeled ID data &  71.6\\
                 ODIN \cite{liang2017enhancing} & Yes & No & 3 & Labeled ID data &  90.7 \\
                 ELOC \cite{vyas132018out} & Yes & Ensemble & 15 & Labeled ID data &  96.3 \\ 
                 Proposed & No & Fine tune & 2 & Labeled ID data \& unlabeled data & 99.6\\
                \bottomrule
            \end{tabular}
        } 	
        \vspace{-15pt}
     \end{table*}
     
   Currently, there are several different methods for out-of-distribution detection. A summary of key methods described are shown in \Tref{tbl:comp}.
   
   As the simplest method, Hendrycks \& Gimpel \cite{hendrycks2016baseline} attempted to detect OOD samples depending on the predicted softmax class probability, which is based on the observation that the prediction probability of incorrect and OOD samples tends to be lower than that of the correct samples. However, they also found that some OOD samples still can be classified overconfidently by pre-trained neural networks, which limits the performance of detection.

    To improve the effectiveness of Hendrycks \& Gimpel's method \cite{hendrycks2016baseline}, Lee et al. \cite{lee2017training} used modified generative adversarial networks \cite{goodfellow2014generative}, which involves training a generator and a classifier simultaneously. They trained the generator to generate `boundary' OOD samples that appear to be at the boundary of the given in-distribution data manifold, while the classifier is encouraged to assign these OOD samples uniform class probabilities in order to generate less confident predictions on them.
    
    Liang et al. \cite{liang2017enhancing} also proposed an improved solution, which applies temperature scaling and input preprocessing, called Out-of-DIstribution Detector for Neural Networks (ODIN). They found that by scaling the unnormalized outputs (logits) before the final softmax layer by a large constant (temperature scaling), the difference between the largest logit and the remaining logits is larger for ID samples than for OOD samples, which shows that the separation of the softmax scores between ID and OOD samples is increased. In addition, if they add some small perturbations to the input through the loss gradient, which increases the maximum predicted softmax score, the increases on ID samples are greater than those on OOD samples. Based on these observations, the authors first scaled the logits at a high temperature value to calibrate softmax scores and then pre-processed the input by perturbing it with the loss gradient to further increase the difference between the maximum softmax scores of ID and OOD samples. Their approach outperforms the baseline method \cite{hendrycks2016baseline}.

    Lee et al. \cite{lee2018simple} and Quintanilha et al. \cite{anonymous2019detecting} extracted low- and upper-level features from DNNs to calculate a confidence score for detecting OOD samples. However, these two methods require 1,000 labeled OOD samples to train a logistic regression detector to achieve stable performance.
	
	Some studies \cite{DD, MyDD, weinshall2012beyond} utilized the hierarchical relations between labels and trained two classifiers to have different generality (a general classifier and a specific classifier) by using different levels of the label hierarchy. OOD samples can be detected by incongruence between the general classifier and the specific classifier, but the requirement of label hierarchy limits the application of these methods.

    There are some other studies on open-set classification \cite{bendale2015towards, bendale2016towards, ge2017gopenmax, rudd2018evm, scheirer2013toward, scheirer2014probability, yoshihashi2019open}, which involve tasks very similar to OOD detection. Bendale \& Boult \cite{bendale2016towards} proposed a new layer called openMax that can calculate the score for an unknown class by taking the weighted average of all other classes obtained from a Weibull distribution.

    The current state-of-the-art method for OOD detection is the ensemble of self-supervised leave-out classifiers proposed by Vyas et al. \cite{vyas132018out}. They divided the training ID data into $K$ partitions and assign one partition as OOD and the remaining partitions as ID to train $K$ classifiers by a novel loss function, called margin entropy loss, to increase the prediction confidence of ID samples and decrease the prediction confidence of OOD samples. During test time, they used an ensemble of these $K$ classifiers for detecting OOD samples in addition to temperature scaling and input pre-processing proposed in ODIN \cite{liang2017enhancing}.

    Compared to previous studies, our method fine-tunes the neural network by utilizing unlabeled data for unsupervised learning. Our unlabeled data is all or a part of test data. 

\section{Method}

    In this section, we present our proposed method for OOD detection. First, we describe the problem statements in \Sref{sec:problem}. Second, we illustrate the overall concept of our method in \Sref{sec:idea}. Then, our loss function is explained in \Sref{sec:loss} and we detail the actual training procedure of our method in \Sref{sec:step}. Finally, we introduce the method used to detect OOD samples at inference time in \Sref{sec:test}.

 \subsection{Problem Statement}

 \label{sec:problem}
 We suppose that an ID image-label pair, $\{\mathbf{x_{in}},y_{in}\}$, drawn from a set of labeled ID images $\{X_{in}, Y_{in}\}$ , is accessible, as well as an unlabeled image, $\mathbf{x_{ul}}$, drawn from unlabeled images $X_{ul}$. The ID sample $\{\mathbf{x_{in}},y_{in}\}$ can be classified into $K$ classes, which means $y_{in} \in K$. Please note that $\mathbf{x_{ul}}$ can be either an ID image or an OOD image and $\exists \{\mathbf{x_{ul}}, y_{ul}\}, y_{ul} \notin K$, so we do not know whether this image is from in- or out-of-distribution. Unlike previous methods, we use $\mathbf{x_{ul}}$ for unsupervised training, which is realistic for real-world applications.

 The goal of our method is to distinguish whether the image $\mathbf{x_{ul}}$ is from in-distribution or not. For this objective, we have to train the network to predict different softmax class probabilities for ID samples and OOD samples. 
 
\subsection{Overall Concept}
\label{sec:idea}
    Hendrycks \& Gimpel \cite{hendrycks2016baseline} showed that the prediction probability of OOD samples tends to be lower than the prediction probability of ID samples; thus, OOD samples are closer to the class boundaries and more likely to be misclassified or classified with low confidence by the classifier learned from ID samples (the upper part of \Fref{fig:overview}).
    
    Based on their findings, we further propose a two-head CNN inspired by \cite{saito2017maximum}, consisting of a feature extractor network, $E$, which takes inputs $\mathbf{x_{in}}$ or $\mathbf{x_{ul}}$, and two classifier networks, $F_1$ and $F_2$, which take features from $E$ and classify them into $K$ classes. Classifier networks $F_1$ and $F_2$ output a $K$-dimensional vector of logits; then, the class probabilities can be calculated by applying the softmax function for the vector. The notation $p_1(\mathbf{y}|\mathbf{x})$ and $p_2(\mathbf{y}|\mathbf{x})$ are used to denote the $K$-dimensional softmax class probabilities for input $\mathbf{x}$ obtained by $F_1$ and $F_2$, respectively. 
   Differing from \cite{saito2017maximum} which aligns the distributions of two datasets for domain adaptation, we train the network on different loss functions with a different training procedure to detect the difference between the distributions of two datasets.
   
   \begin{figure}
        \centering
        \includegraphics[width=8cm]{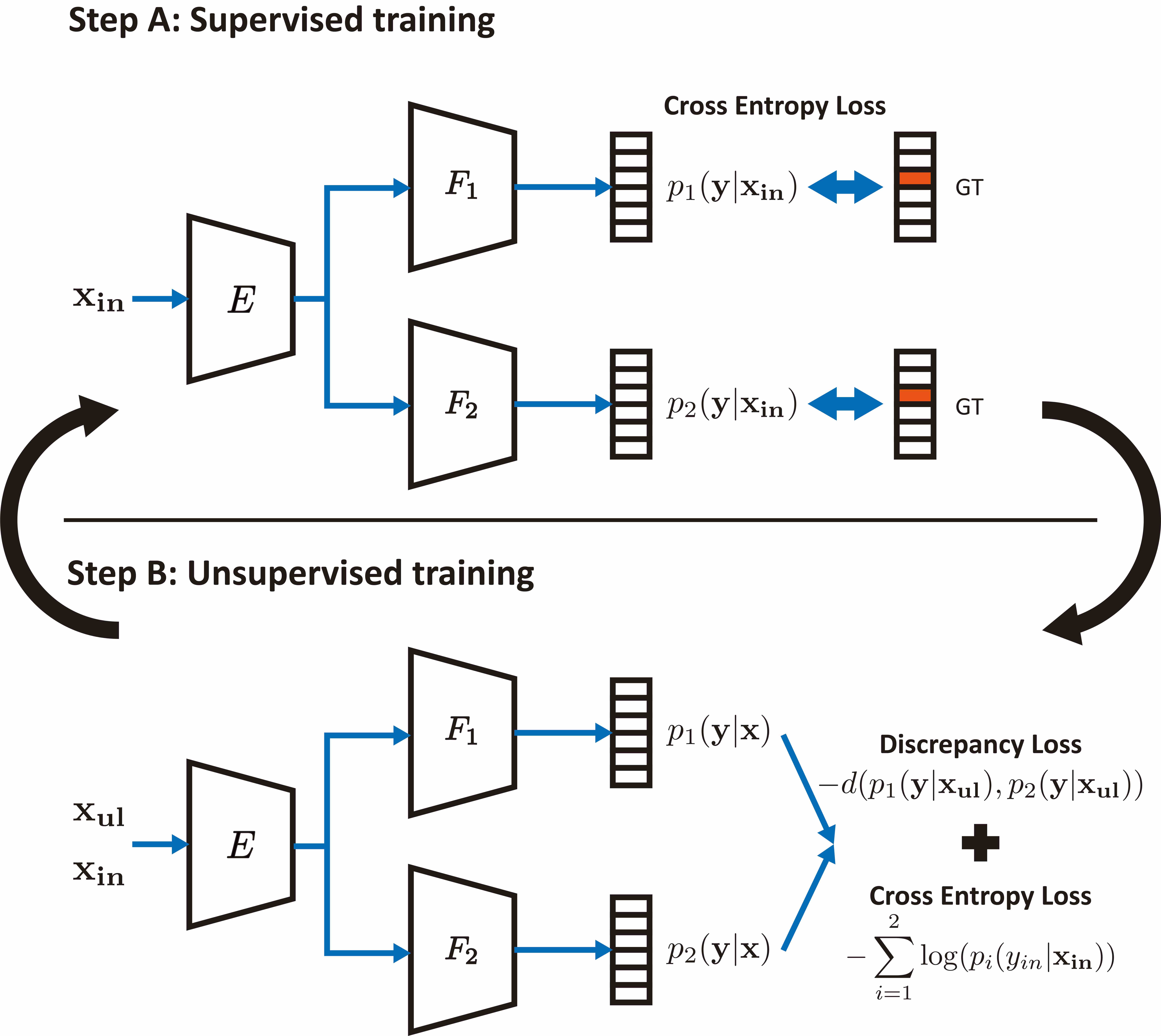}
        \vspace{-10pt}
        \caption{Fine-tuning steps of our method. Our network has one extractor ($E$) and two classifiers ($F_1, F_2$). \textbf{Step A}: Train the network to classify ID samples correctly under supervision. \textbf{Step B}: The classifiers learn to maximize the discrepancy in an unsupervised manner, which helps to detect OOD samples.}
        \label{fig:method}
        \vspace{-15pt}
    \end{figure}

\begin{figure}[t]
    \centering
    \includegraphics[keepaspectratio, width=8.5cm]
    {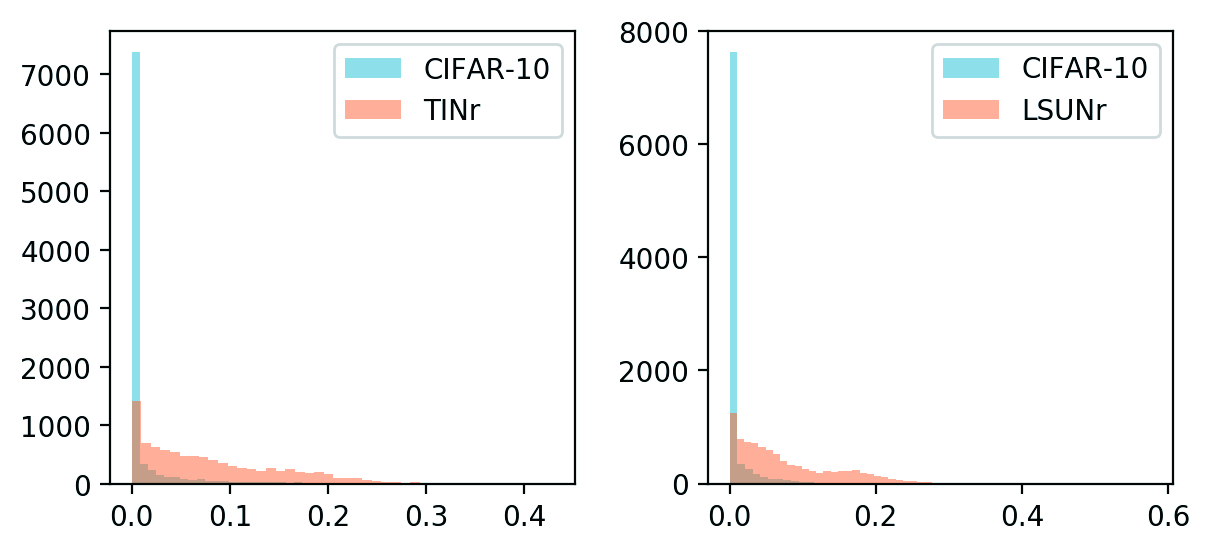}
    \vspace{-25pt}
    \caption{Histogram of the discrepancy (L1 distance) between the two classifiers trained on ID samples.}\label{fig:before}
    \vspace{-15pt}
  \end{figure}

    We found that when the two classifiers ($F_1$ and $F_2$) are initialized with random initial parameters and then trained on ID samples supervisedly, they will have different characteristics and classify OOD samples differently (the lower part side of \Fref{fig:overview}).
    \Fref{fig:before} shows the disagreement (L1 distance) between the two classifiers' outputs of unlabeled ID (CIFAR-10) and OOD (TinyImageNet-resized and LSUN-resized) samples after training the network on labeled ID samples supervisedly. We can confirm that most OOD samples have larger discrepancy than ID samples in \Fref{fig:before}.
    
	By utilizing this characteristic, if we can measure the disagreement between the two classifiers and train the network to maximize this disagreement, the network will push OOD samples outside the manifold of ID samples. Discrepancy, $d(p_1(\mathbf{y}|\mathbf{x}), p_2(\mathbf{y}|\mathbf{x}))$, is introduced to measure the divergence between the two softmax class probabilities for an input. Consequently, we can separate OOD samples and ID samples according to the discrepancy between the two classifiers' outputs.
    
\subsection{Discrepancy Loss}
\label{sec:loss}
    We define the discrepancy loss as the following equation:
    \begin{equation}
        \label{eq:dis}
        d(p_1(\mathbf{y}|\mathbf{x}), p_2(\mathbf{y}|\mathbf{x})) \\= H(p_1(\mathbf{y}|\mathbf{x})) - H(p_2(\mathbf{y}|\mathbf{x})),
    \end{equation}
    where $H(\cdot)$ is the entropy over the softmax distribution.

    When the network is trained to maximize this discrepancy term, it maximizes the entropy of $F_1$'s output, which encourages $F_1$ to predict equal probabilities of all the classes, and minimizes the entropy of $F_2$'s output, which encourages $F_2$ to predict high probability of one class simultaneously. Since OOD samples are outside the support of the ID samples, the discrepancy between the two classifiers' outputs of OOD samples will be larger. This is demonstrated empirically in \Sref{sec:exp}.

\subsection{Training Procedure}
\label{sec:step}
    
    As the previous discussion in \Sref{sec:idea}, we need to train our network to classify ID samples correctly and maximize $d(p_1(\mathbf{y}|\mathbf{x}), p_2(\mathbf{y}|\mathbf{x}))$ at the same time. To achieve this, we propose a training procedure consisting of one pre-training step and two repeating fine-tuning steps. Pre-training step uses labeled ID samples $\{X_{in}, Y_{in}\}$ to train the classifier. Then, both $\{X_{in}, Y_{in}\}$ and unlabeled samples $X_{ul}$ are used to train the network for separating ID and OOD samples while keeping correct classification of ID samples in fine-tuning steps.
    In principle, we use the test data as the unlabeled data. In addition, the unlabeled data can be only a part of the test data. In the ablation studies in \Sref{sec:ablation} , we conduct experiments in the cases of varying sizes and types of unlabeled data. 
    
    \textbf{Pre-training}: First, we train the network to learn discriminative features and classify the ID samples correctly under the supervision of labeled ID samples. The network is trained to minimize the cross entropy with the following loss:
    \begin{equation}
        \label{eq:sf}
        \mathcal{L}_{sup} = - \frac{1}{|X_{in}|} \sum_{\mathbf{x_{in}} \in X_{in}}\sum_{i=1}^{2}\log(p_i(y_{in}|\mathbf{x_{in}})).
    \end{equation}

    \textbf{Fine-tuning}: Once the network converges, we start to fine-tune the network to detect OOD samples by repeating the following two steps at mini-batch level.
    \begin{itemize}
    \vspace{-5pt}
        \setlength\itemsep{0pt}
        \item \textbf{Step A} First, during the fine-tuning process, we keep training the network to classify the labeled ID samples correctly by supervised learning (Step A in \Fref{fig:method}) with \Eref{eq:sf} to maintain the manifold of ID samples. This step is helpful to improve the performance of our algorithm.
        \item \textbf{Step B} Then, we train the network to increase the discrepancy in an unsupervised manner in order to make the network detect the OOD samples that do not have the support of the ID samples (Step B in \Fref{fig:method}). In this step, we also use labeled ID samples to reshape the support. We add classification loss on the labeled ID samples. The same mini-batch size of labeled and unlabeled samples are utilized to update the model at this step. As a result, we train the network to minimize the following loss:
    \end{itemize}
        \begin{equation}
            \mathcal{L} = \mathcal{L}_{sup} + \mathcal{L}_{unsup}\label{eq:loss}
        \end{equation}
        \begin{equation}
        \mathcal{L}_{sup} = - \frac{1}{|X_{in}|} \sum_{\mathbf{x_{in}} \in X_{in}}\sum_{i=1}^{2}\log(p_i(y_{in}|\mathbf{x_{in}}))
        \end{equation}
        \begin{equation}
            \mathcal{L}_{unsup} = \max{(m-\frac{\displaystyle \sum\limits_{\mathbf{x_{ul}} \in X_{ul}}d(p_1(\mathbf{y}|\mathbf{x_{ul}}), p_2(\mathbf{y}|\mathbf{x_{ul}}))}{|X_{ul}|}, 0)}.\label{eq:unsuploss}
        \end{equation}
        \begin{itemize}
        \item[] If the average discrepancy of unlabeled samples is greater than the margin $m$, the unsupervised loss will equal its minimum value, zero; thus, the margin $m$ is helpful for preventing overfitting.
        \end{itemize}

\subsection{Inference}
\label{sec:test}
    At inference time, in order to distinguish between in- and out-of-distribution samples, a straightforward solution would be to use the discrepancy defined in \Sref{sec:loss}, but this term does not include the discrepancy of each class. We consider the L1 distance between the two classifiers' outputs. When the distance is above a detection threshold $\delta$, we assign the sample as an out-of-distribution sample, denoted by
    \begin{equation}
        \sum_{i=1}^{K}|p_1(y_i|\mathbf{x}) - p_2(y_i|\mathbf{x})| > \delta.
    \end{equation}

\section{Experiments}
\label{sec:exp}
	In this section, we discuss our experimental settings and results. We describe a diverse set of in- and out-of-distribution dataset pairs, neural network architectures and evaluation metrics. We also demonstrate the effectiveness of our method by comparing it against the current state-of-the-art methods, resulting in our method significantly outperforming them. We ran all experiments using PyTorch 0.4.1 \cite{paszke2017automatic}.
	
    \subsection{OOD Detection on benchmarks}
    As the benchmarks of OOD detection, ODIN \cite{liang2017enhancing} and Ensemble of Leave-Out Classifiers (ELOC) \cite{vyas132018out} introduced several benchmark datasets and evaluation metrics to evaluate the performance of OOD detectors.
	\subsubsection{Neural Network Architecture}
	\label{sec:nn}
	Following \cite{liang2017enhancing, vyas132018out}, we implemented our network based on two state-of-the-art neural network architectures, DenseNet \cite{huang2017densely} and Wide ResNet (WRN) \cite{Zagoruyko2016WRN}. We used the modules of DenseNet/Wide ResNet until the {\it average-pooling} layer just before the last {\it full-connected} layer as the extractor, and one {\it full-connected} layer as the classifier.
    
	In the pre-training step proposed in \Sref{sec:step}, we used stochastic gradient descent (SGD) to train DenseNet-BC for 300 epochs and Wide ResNet for 200 epochs. The learning rate started at 0.1 and dropped by a factor of 10 at 50\% and 75\% of the training progress, respectively. After the pre-training step, we further fine-tuned the network in the fine-tuning steps proposed in \Sref{sec:step} for 10 epochs with learning rate 0.1, margin $m=1.2$ to detect OOD samples.

    Furthermore, for fair comparison, we used two classifiers and calculated the average score of these two classifiers as final output in the other methods, ODIN \cite{liang2017enhancing} and Ensemble of Leave-Out Classifiers \cite{vyas132018out}, since our method has two classifiers which resulted in a few more parameters.
	
	\subsubsection{In-Distribution}
    CIFAR-10 (contains 10 classes) and CIFAR-100 (contains 100 classes) \cite{Krizhevsky09learningmultiple} datasets were used as in-distribution datasets to train deep neural networks for image classification. They both consist of 50,000 images for training and 10,000 images for testing, with the image size of $32 \times 32$. The images in the train split were used as $X_{in}$ in our experiment.
    
    \begin{table*}[t]
        \centering
        \caption{The result of distinguishing in- and out-of-distribution test set data on OOD benchmarks. Our method is compared with ODIN \cite{liang2017enhancing} and Ensemble of Leave-Out Classifiers (ELOC) \cite{vyas132018out}. As described in \Sref{sec:nn}, ODIN \cite{liang2017enhancing} and ELOC \cite{vyas132018out} are modified to have two ensembled {\it full-connected} layers for fair comparison. $\uparrow$ indicates larger value is better, and $\downarrow$ indicates lower value is better. All values are percentages.}
        \vspace{-5pt} 
		\label{tbl:OOD_result}
        \scalebox{0.905}{
            \tabcolsep = 1.3mm
            \begin{tabular}{cc|ccc|ccc|ccc|ccc|ccc}
                \toprule
                 & \begin{tabular}{c}OOD \\ dataset\end{tabular} & \multicolumn{3}{c|}{\begin{tabular}{c}FPR \\ (95\% TPR)\\$\downarrow$\end{tabular}} & \multicolumn{3}{c|}{\begin{tabular}{c}Detection \\ Error\\$\downarrow$\end{tabular}} & \multicolumn{3}{c|}{\begin{tabular}{c}AUROC\\$\uparrow$\end{tabular}} & \multicolumn{3}{c|}{\begin{tabular}{c}AUPR In\\$\uparrow$\end{tabular}} & \multicolumn{3}{c}{\begin{tabular}{c}AUPR Out\\$\uparrow$\end{tabular}} \\ \hline
                & & ODIN & ELOC & Ours  & ODIN & ELOC &  \multicolumn{1}{c|}{Ours}  & ODIN & ELOC  &  \multicolumn{1}{c|}{Ours} & ODIN & ELOC &  \multicolumn{1}{c|}{Ours}  &  ODIN & ELOC  &  \multicolumn{1}{c}{Ours} \\ \cline{3-17} 
                \multirow{5}{*}{\begin{sideways}\begin{tabular}{c}\textbf{Dense-BC} \\ CIFAR-10\end{tabular}\end{sideways}} & TINc & 3.6 & 1.5 & \textbf{0.1} & 4.2 & 3.0 & \textbf{0.7} & 99.2 & 99.6 & \textbf{99.9} & 99.2 & 99.6 & \textbf{100.0} & 99.2 & 99.6 & \textbf{99.9}  \\
                & TINr  &  10.1   &  3.2  & \textbf{1.7}  & 6.9 & 4.0 &  \textbf{2.3}   & 98.2 & 99.3 & \textbf{99.6} & 98.3 & 99.3 &  \textbf{99.6}   & 98.1 & 99.2 & \textbf{99.6}  \\
                & LSUNc &  6.0   &  3.4  & \textbf{0.2}  & 5.3 & 4.1 &\textbf{0.7}     & 98.7 & 99.3 &\textbf{99.9}     & 98.7 & 99.3 &\textbf{99.9}     & 98.6 & 99.3 & \textbf{99.9}   \\ 
                & LSUNr &  3.5   &  1.4  & \textbf{0.4}  & 4.2 & 2.7 &  \textbf{1.1}   & 99.2 & 99.6 & \textbf{99.9}    & 99.3 & 99.6 & \textbf{99.9}    & 99.2 & 99.6 & \textbf{99.9}  \\
                & iSUN  &  5.9   &   -  & \textbf{0.6}  & 5.3 & - &  \textbf{1.3}   & 98.9 & - & \textbf{99.9}    & 99.0 & - &  \textbf{99.9}   & 98.9 & - &  \textbf{99.9} \\ \hline
                \multirow{5}{*}{\begin{sideways}\begin{tabular}{c}\textbf{Dense-BC} \\ CIFAR-100\end{tabular}\end{sideways}} & TINc & 20.6 & 8.8 & \textbf{0.2} & 10.2 & 6.6 &  \textbf{0.7}   & 96.4 & 98.3 &   \textbf{99.9}  & 96.7 & 98.4 & \textbf{99.9}    & 96.1 & 98.3 &  \textbf{99.9} \\
                & TINr  &  43.1   &  20.6  & \textbf{1.9}  & 17.2 & 10.2 &  \textbf{2.0}   & 90.7 & 96.2 &   \textbf{99.6}  & 91.0 & 96.5 &  \textbf{99.6}   & 89.8 & 96.0 &  \textbf{99.7} \\
                & LSUNc &  21.9   &  16.2 & \textbf{0.3}  & 10.1 & 9.3 &  \textbf{0.6}   & 95.9 & 97.0 &  \textbf{99.9}   & 96.4 & 97.3 &  \textbf{99.9}   & 96.0 & 96.8 & \textbf{99.9}  \\\
                & LSUNr &  43.2   &  13.1 & \textbf{0.4}  & 24.5 & 7.7 &  \textbf{0.6}   & 91.0 & 97.6 &  \textbf{99.9}   & 91.5 & 97.9 &  \textbf{99.9}   & 89.8 & 97.3 &  \textbf{99.9} \\
                & iSUN  &  45.4   &  - & \textbf{1.3}  & 17.2 & - &  \textbf{1.6}   & 90.5 & - &   \textbf{99.7}  & 90.9 & - &  \textbf{99.6}   & 89.1 & - & \textbf{99.7}  \\
                \midrule
                \multirow{5}{*}{\begin{sideways}\begin{tabular}{c}\textbf{WRN-28-10} \\ CIFAR-10\end{tabular}\end{sideways}} & TINc & 16.6 & 1.5 & \textbf{0.2} & 8.9 & 3.0 & \textbf{0.6} & 96.9 & 99.6 & \textbf{100.0} & 97.3 & 99.6 & \textbf{100.0} & 96.5 & 99.6 & \textbf{100.0}  \\
                & TINr  &  6.1   &  5.5  & \textbf{0.8}  & 5.5 & 5.1 &  \textbf{1.8}   & 98.8 & 98.9 &  \textbf{99.7}   & 98.9 & 99.0 &  \textbf{99.7}   & 98.8 & 98.8 &  \textbf{99.7} \\
                & LSUNc &  20.3   &  1.6  & \textbf{0.0}  & 9.6 & 3.0 &  \textbf{0.2}   & 96.4 & 99.6 &  \textbf{100.0}   & 96.9 & 99.6 &  \textbf{100.0}   & 95.8 & 99.5 &  \textbf{100.0} \\
                & LSUNr &  4.6   &  0.9  & \textbf{0.4}  & 4.7 & 2.5 &  \textbf{1.7}   & 99.0 & 99.7 &  \textbf{99.8}   & 99.1 & 99.7 &  \textbf{99.8}   & 99.0 & 99.7 &  \textbf{99.8} \\
                & iSUN  &  3.7   &   -  & \textbf{0.3}  & 4.3 & - &  \textbf{1.1}   & 99.2 & - &   \textbf{99.9}  & 99.3 & - &  \textbf{99.9}   & 99.2 & - & \textbf{99.9}  \\ \hline
                \multirow{5}{*}{\begin{sideways}\begin{tabular}{c}\textbf{WRN-28-10} \\ CIFAR-100\end{tabular}\end{sideways}} & TINc & 33.3 & 8.6 & \textbf{0.6}  & 13.4 & 6.3 &  \textbf{1.6}   & 93.9 & 98.5 & \textbf{99.8}    & 94.6 & 98.6 &   \textbf{99.7}  & 92.8 & 98.4 &  \textbf{99.8} \\
                & TINr  &  35.8   &  18.9  &  \textbf{1.6} & 15.4 & 9.1 &  \textbf{2.3}   & 92.7 & 96.8 &  \textbf{99.6}   & 93.2 & 97.1 &  \textbf{99.5}   & 92.2 & 96.4 &  \textbf{99.6} \\
                & LSUNc &  34.9   &  25.1 &  \textbf{0.5} & 15.5 & 10.6 &  \textbf{1.4}   & 92.7 & 96.0 &   \textbf{99.8}  & 93.1 & 96.5 &   \textbf{99.8}  & 92.5 & 95.5 & \textbf{99.8}  \\
                & LSUNr &  34.9   &  12.8 & \textbf{0.6}  & 14.9 & 7.4 &  \textbf{1.4}   & 93.1 & 97.6 &  \textbf{99.7}   & 93.6 & 97.8 &  \textbf{99.6}   & 92.8 & 97.4 &  \textbf{99.8} \\
                & iSUN  &  34.1   &  - &  \textbf{0.9} & 14.6 & - &   \textbf{1.4}  & 93.3 & - &  \textbf{99.6}   & 93.9 & - &  \textbf{99.5}   & 92.5 & - & \textbf{99.7}  \\
                \bottomrule
            \end{tabular}
        } 	
        \vspace{-15pt}
     \end{table*}
	
	\subsubsection{Out-of-Distribution}
	We followed the benchmarks given in \cite{liang2017enhancing, vyas132018out} and used the OOD datasets below in our experiments:
    \begin{enumerate}
        \setlength\itemsep{-3pt}
        \item \textbf{TinyImageNet (TIN).}  The Tiny ImageNet dataset \cite{imagenet_cvpr09} contains 10,000 test images from 200 different classes, which are drawn from the original 1,000 classes of ImageNet \cite{imagenet_cvpr09}. TinyImageNet-crop (TINc) and TinyImageNet-resize (TINr) are constructed by either randomly cropping or downsampling each image to a size of $32 \times 32$.
        \item \textbf{LSUN.}  The Large-scale Scene Understanding dataset (LSUN) consists of 10,000 test images from 10 different scene categories.\cite{song2015construction}. Similar to TinyImageNet, by randomly cropping and downsampling the LSUN test set, two datasets LSUN-crop (LSUNc) and LSUN-resize (LSUNr) are constructed.
        \item \textbf{iSUN.} The iSUN is a subset of SUN \cite{xiao2010sun}, which is used for gaze tracking, deployed on Amazon Mechanical Turk using a webcam \cite{xu2015turkergaze}. It contains 8,925 scene images, and all images are downsampled to a size of $32 \times 32$.
    \end{enumerate}
    For each in-distribution dataset (test split) and each out-of-distribution dataset, 1,000 images (labeled as ID or OOD) were randomly held out for validation, such as parameter tuning and early stopping, while the remaining test images containing unlabeled ID or OOD samples were used as $X_{ul}$ for unsupervised training and evaluation. These datasets are provided as a part of ODIN \cite{liang2017enhancing} code release\footnote{\url{github.com/facebookresearch/odin}}.

	\subsubsection{Evaluation Metrics}
	\label{sec:metrics}
    We followed the same metrics used by \cite{liang2017enhancing, vyas132018out} to measure the effectiveness of our method in distinguishing between in- and out-of-distribution samples. TP, TN, FP, FN are used to denote true positives, true negatives, false positives, and false negatives, respectively.
    \begin{enumerate}
        \setlength\itemsep{-3pt}
        \item \textbf{FPR at 95\% TPR} shows the false positive rate (FPR) at 95\% true positive rate (TPR). True positive rate can be computed by TPR = TP / (TP+FN), while the false positive rate (FPR) can be computed by FPR = FP / (FP+TN).
        \item \textbf{Detection Error} measures the minimum misclassification probability, which is calculated by the minimum average of false positive rate (FPR) and false negative rate (FNR) over all possible score thresholds.
        \item \textbf{AUROC} is the Area Under the Receiver Operating Characteristic curve and can be calculated by the area under the FPR against TPR curve.
        \item \textbf{AUPR In} is the Area Under the Precision-Recall curve and can be calculated by the area under the precision = TP / (TP+FP) against the recall = TP / (TP+FN) curve. For AUPR In, in-distribution images are specified as positive.
        \item \textbf{AUPR Out} is similar to the metric AUPR-In. The difference is that out-of-distribution images are specified as positive in AUPR Out.
    \end{enumerate}
    \vspace{-13pt}
    
    \subsubsection{Experimental Results}
    The results are summarized in \Tref{tbl:OOD_result}, which shows the comparison of our method, ODIN \cite{liang2017enhancing} and Ensemble of Leave-Out Classifiers (ELOC) \cite{vyas132018out} on various benchmarks. In addition, ELOC \cite{vyas132018out} does not have results for iSUN as an OOD dataset because they use the whole iSUN as a validation dataset. We implemented two ensembled {\it full-connected} layers in ODIN \cite{liang2017enhancing} and ELOC \cite{vyas132018out}, and their performance is almost the same as that of the single classifier (one {\it full-connected} layer) in their original papers.

    \Tref{tbl:OOD_result} clearly shows that our approach significantly outperforms other existing methods, including ODIN \cite{liang2017enhancing} and ELOC \cite{vyas132018out} (which is the ensemble of five models), across all neural network architectures on all of the dataset pairs.

    TinyImageNet-resize, LSUN-resize and iSUN, which contain the images with full objects as opposed to the cropped parts of objects, are considered more difficult to detect. Our proposal shows highly accurate results on these more challenging datasets.
    
        \begin{figure}[t]
		\centering
        \begin{subfigure}[b]{1\linewidth}
        \centering
            \includegraphics[keepaspectratio, width=8.5cm]
            {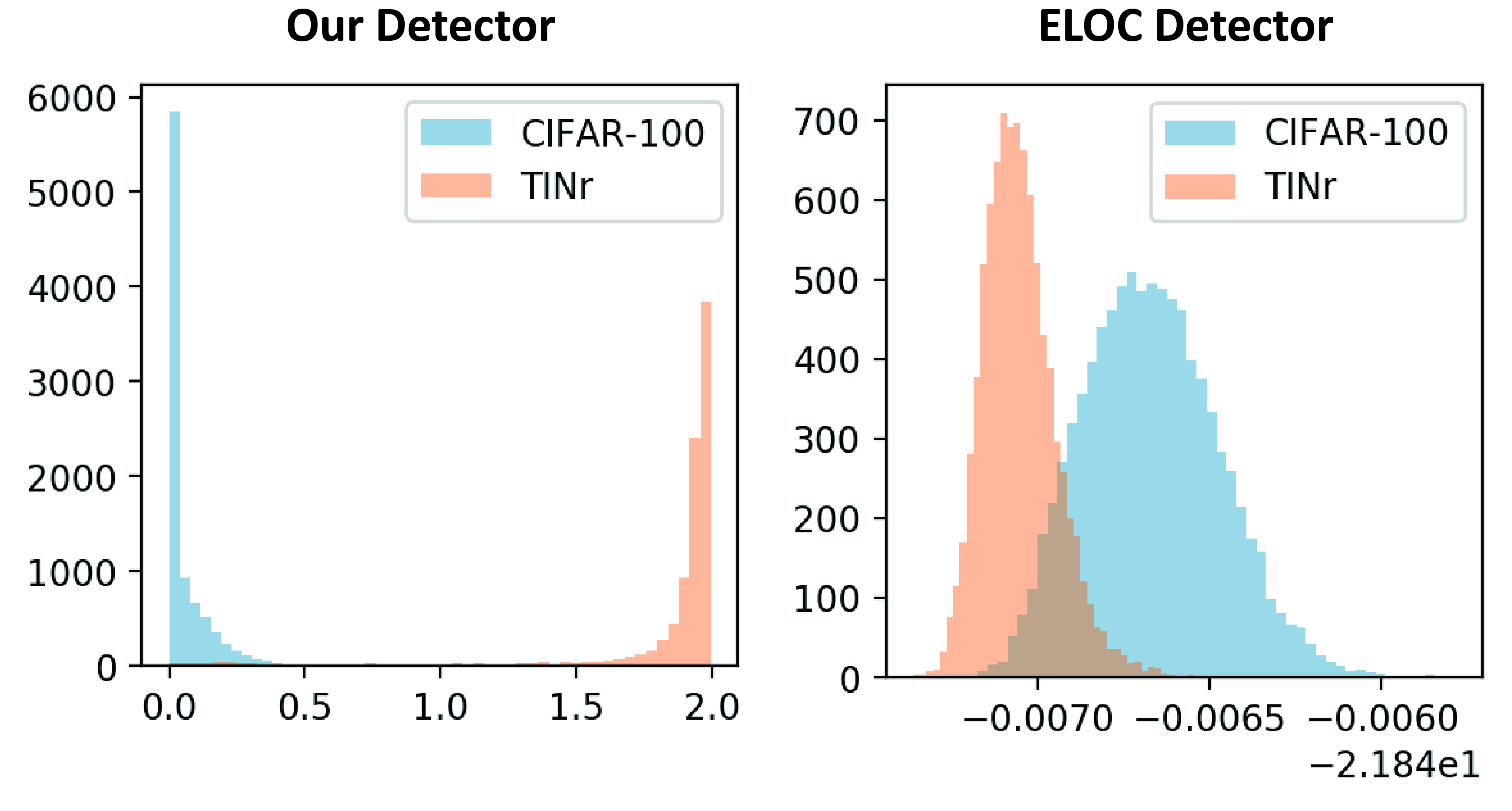}
            \vspace{-20pt}
            \subcaption{Histogram of ID and OOD detection scores of the proposed method and ELOC \cite{vyas132018out}.}\label{fig:score}
        \end{subfigure}
        ~
        \begin{subfigure}[b]{1\linewidth}
            \centering
            \includegraphics[keepaspectratio, width=8.5cm] 
            {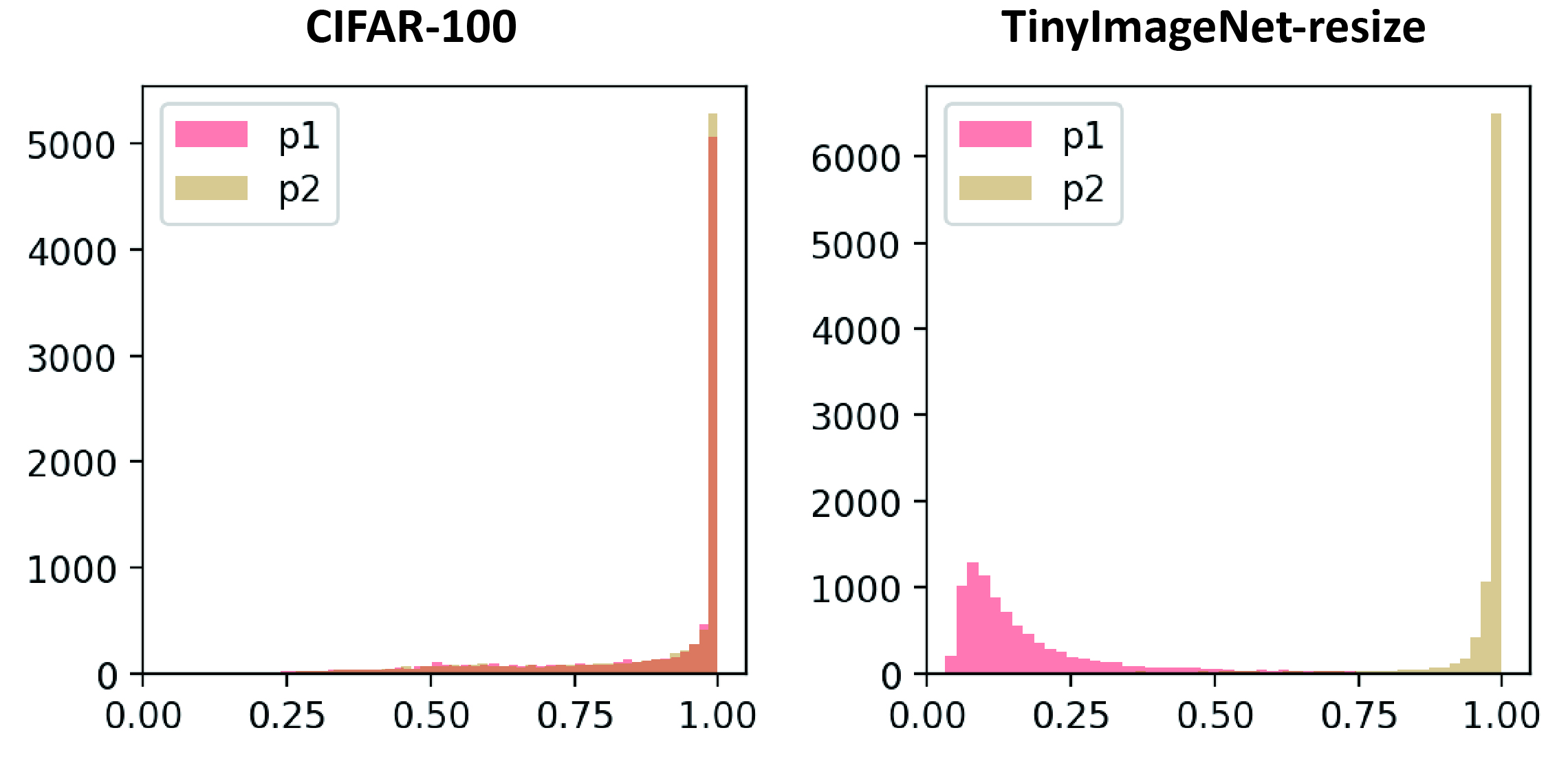} 
            \vspace{-20pt} 
            \subcaption{Histogram of the two classifiers' maximum softmax scores after the fine-tuning.}\label{fig:dis}
        \end{subfigure} 
        \vspace{-20pt} 
        \caption{The visualization of the result.}\label{fig:vis}
        
        \vspace{-15pt} 
      \end{figure}   

    Noticeably, our method is very close to distinguishing between in- and out-of-distribution samples perfectly on most dataset pairs. As shown in \Fref{fig:score}, we compared our OOD detector based on DenseNet-BC with ELOC \cite{vyas132018out} when in-distribution data is CIFAR-100 and out-of-distribution data is TinyImageNet-resize. These figures show that the proposed method has much less overlap between OOD samples and ID samples on all the dataset pairs compared to ELOC \cite{vyas132018out}, indicating that our method separates ID and OOD samples very well.
    
    Another merit of our method is that we can use a simple threshold $1.0$ to separate ID and OOD samples as shown in \Fref{fig:score}. On the other hand, it is very difficult to decide an interpretable threshold of that value in ELOC.

    We also plot the histogram of the two classifiers' maximum softmax scores of $X_{ul}$ in \Fref{fig:dis}.
    \Fref{fig:dis} shows that the distribution of $X_{ul}$'s $p_1$ and $p_2$ differs significantly according to whether the sample is ID (CIFAR-100) or not (TINr) after the fine-tuning. For convenience, we use $p_{1k}$ and $p_{2k}$ to denote the probability output of $p_1$ and $p_2$ for class $k$, respectively. The discrepancy loss makes OOD samples' $\max_{k}{p_{1k}}$ close to $1/K$ and $\max_{k}{p_{2k}}$ close to 1. On the other hand, ID samples' $\max_{k}{p_{1k}}$ are almost the same as $\max_{k}{p_{2k}}$ due to the support we added to ID samples in Step A and \Eref{eq:loss} in Step B. 
\vspace{-5pt}
\subsubsection{Limitation}
\vspace{-5pt}
    Since our method needs to fine-tune the classifier to detect OOD samples which changes the decision boundary, we observed a 5\% drop of classification accuracy compared to the original classifier before the fine-tuning. This problem could be solved by using the original classifier to classify ID samples with some runtime increase and it is still much more acceptable than ELOC \cite{vyas132018out} using an ensemble of five models which needs more runtime and computing resources.

\begin{table}[t]
    \caption{Results of ablation studies on CIFAR-100 as ID and TinyImageNet-crop as OOD.}
    \vspace{-5pt}
    \centering
    \scalebox{1.03}{
    \begin{tabular}{ccc|cccc}\hline
      \multicolumn{3}{c|}{\#ID in $X_{ul}$}&9k&5k&2k&1k\\ \hline
      \multicolumn{3}{c|}{\#OOD in $X_{ul}$}&9k&5k&2k&1k\\ \hline\hline
      \multicolumn{3}{c|}{Detection Error (\%)}& 0.7 & 0.5 & 0.9 & 1.5 \\ \hline
      \multicolumn{3}{c|}{ID Discrepancy Loss}& 0.05 & 0.06 & 0.08 & 0.05 \\ \hline
      \multicolumn{3}{c|}{OOD Discrepancy Loss}& 3.08 & 3.05 & 2.84 & 2.68 \\ \hline
      \multicolumn{3}{c}{}\\\hline
      \multicolumn{3}{c|}{\#ID in $X_{ul}$}&9k&5k&9k&9k\\ \hline
      \multicolumn{3}{c|}{\#OOD in $X_{ul}$}&2k&2k&1k&500\\ \hline\hline
      \multicolumn{3}{c|}{Detection Error (\%)}& 0.3 & 0.4& 1.2 & 3.8 \\ \hline
      \multicolumn{3}{c|}{ID Discrepancy Loss}& 0.62 & 0.26 & 1.05 & 1.08 \\ \hline
      \multicolumn{3}{c|}{OOD Discrepancy Loss}& 4.03 & 3.90 & 3.93 & 3.40 \\ \hline
    \end{tabular}
    }
  \label{tbl:ablation}
  \vspace{-10pt}
\end{table}

\begin{table}[t]
  \caption{Results of ablation studies on CIFAR-100 as ID and other datasets as OOD.}
  \vspace{-5pt}
  \centering
  \scalebox{0.95}{
  \tabcolsep = 1.3mm
  \begin{tabular}{ccc|ccc}\hline
    \multicolumn{3}{c|}{OOD in $X_{ul}$}&TINc+LSUNc&TINc&LSUNc\\ \hline
    \multicolumn{3}{c|}{OOD for testing}&TINc+LSUNc&LSUNc&TINc\\ \hline\hline
    \multicolumn{3}{c|}{Detection Error (\%)}& 0.2 & 0.5 & 0.7 \\ \hline
    \multicolumn{3}{c|}{ID Discrepancy Loss}& 0.03 & 0.05 & 0.04 \\ \hline
    \multicolumn{3}{c|}{OOD Discrepancy Loss}& 3.53 & 3.20 & 2.39 \\ \hline
  \end{tabular}
  }
  \label{tbl:ablation2}
  \vspace{-15pt}
\end{table}

\begin{table*}[t]
    \centering
    \caption{The result of distinguishing in- and out-of-distribution test set data on real-world simulation. Our method is compared with ODIN \cite{liang2017enhancing} and ELOC \cite{vyas132018out}. $\uparrow$ indicates larger value is better, and $\downarrow$ indicates lower value is better. All values are percentages.} 
    \vspace{-5pt}
    \label{tbl:real_result}
    
    \scalebox{0.835}{
        \tabcolsep = 1.1mm
        \begin{tabular}{c|c|ccc|ccc|ccc|ccc|ccc}
            \toprule
             \begin{tabular}{c}ID \\ dataset\end{tabular} & \begin{tabular}{c}OOD \\ dataset\end{tabular} & \multicolumn{3}{c|}{\begin{tabular}{c}FPR \\ (95\% TPR)\\$\downarrow$\end{tabular}} & \multicolumn{3}{c|}{\begin{tabular}{c}Detection \\ Error\\$\downarrow$\end{tabular}} & \multicolumn{3}{c|}{\begin{tabular}{c}AUROC\\$\uparrow$\end{tabular}} & \multicolumn{3}{c|}{\begin{tabular}{c}AUPR In\\$\uparrow$\end{tabular}} & \multicolumn{3}{c}{\begin{tabular}{c}AUPR Out\\$\uparrow$\end{tabular}} \\ \hline
            & & ODIN  & ELOC  & Ours  &  ODIN  & ELOC  &  \multicolumn{1}{c|}{Ours}  &  ODIN  & ELOC  &  \multicolumn{1}{c|}{Ours} &  ODIN  & ELOC  &  \multicolumn{1}{c|}{Ours}  &  ODIN  & ELOC  &  \multicolumn{1}{c}{Ours} \\ \cline{3-17}
            
            \multirow{3}{*}{\begin{tabular}{c}Food \\ (FOOD-101)\end{tabular}} & \begin{tabular}{c}Non Food\\ (TINc)\end{tabular} &  48.2  &  36.7   & \textbf{0.1} & 22.4 & 16.5  &   \textbf{0.2} & 85.3 & 91.5 &  \textbf{100.0} & 92.2  & 91.3 &   \textbf{100.0} & 76.1  & 91.9  &  \textbf{100.0}  \\
            \cline{2-17} 
            & \begin{tabular}{c}Non Food\\ (LSUNc)\end{tabular} & 30.6 & 15.9 & \textbf{0.1} &  16.2 &  10.1 & \textbf{0.2} & 90.9 & 96.3 &  \textbf{100.0} & 95.0  &  95.7 &   \textbf{100.0} & 86.0  &  96.8 &  \textbf{100.0} \\
             \hline
                \multirow{3}{*}{\begin{tabular}{c}Fashion \\ (DeepFashion)\end{tabular}} & \begin{tabular}{c}Non Fashion\\ (TINc)\end{tabular} &   57.7  &  6.1   & \textbf{0.4} & 21.3 & 5.2  &   \textbf{1.1} & 86.4 & 98.7 &  \textbf{99.9} & 95.9  & 98.9 &   \textbf{100.0} & 62.5  & 98.5  &  \textbf{99.7}  \\
            \cline{2-17} 
             & \begin{tabular}{c}Non Fashion\\ (LSUNc)\end{tabular} &  35.8 & 3.6  & \textbf{0.2} & 14.8 & 4.2  &   \textbf{0.7} & 92.9 & 99.1 &  \textbf{99.9} & 98.0  &  99.3 &   \textbf{100.0} & 78.9  &  99.0 &  \textbf{99.8} \\
            \bottomrule
        \end{tabular}
    }
 \vspace{-15pt}
\end{table*}
\vspace{-5pt}
\subsubsection{Ablation Studies}
\vspace{-5pt}
\label{sec:ablation}
Since our approach accesses to unlabeled data $X_{ul}$, we further analyzed the effects of the following factors:

\textbf{The size and the data balance of $X_{ul}$.} We used CIFAR-100 as ID and TinyImageNet-crop as OOD and we changed the number of ID and OOD samples in $X_{ul}$ for unsupervised training.
The result are summarized in \Tref{tbl:ablation}, which shows that our proposed method works under various $X_{ul}$ settings.
Even when 9,000 ID samples and 500 OOD samples are included in $X_{ul}$, our method still have better performance than \cite{liang2017enhancing, vyas132018out}, which means our method is robust to the size of $X_{ul}$ and the percentage of OOD data in $X_{ul}$.
\textit{Please notice that we used all 9,000 ID samples and 9,000 OOD samples for testing, which means totally unseen samples were included during evaluation.}  

\textbf{The selection of OOD data in $X_{ul}$.} To show the effectiveness of our method, we also tried various pairs of OOD datasets for unsupervised training and evaluation.
\Tref{tbl:ablation2} shows that our method still works when multiple datasets are used as OOD or even when OOD dataset used for unsupervised training is different from the OOD dataset for evaluation.

\textbf{The relationship between the discrepancy loss and the detection error.} The mean discrepancy loss in \Eref{eq:dis} of ID and OOD samples in test dataset is shown in \Tref{tbl:ablation} and \Tref{tbl:ablation2}, respectively.
These results show that the discrepancy loss of ID samples is smaller than OOD samples in all settings.
The detection error is lower when the difference between the discrepancy loss of ID and OOD samples is larger, which means ID and OOD samples can be separated by the divergence between the two classifiers' outputs.
     
\vspace{-2pt}
\subsection{OOD Detection on real-world simulation}
\vspace{-2pt}
Since our goal is to benefit applications in real world, we also evaluated our method in two cases of real-world simulation to demonstrate the effectiveness of our method. 
\vspace{-6pt}
\subsubsection{Neural Network Architecture}
\vspace{-6pt}
As previous experiments, we used and pre-trained the same DenseNet-BC \cite{huang2017densely} as \Sref{sec:nn}. We further fine-tuned the network in the fine-tuning steps proposed in \Sref{sec:step} for 10 epochs with a learning rate of 0.1 and margin $m=1.2$ to detect OOD samples.
\vspace{-6pt}
\subsubsection{Real-world Simulation Datasets}
\vspace{-6pt}
Considering domain specific applications, we evaluated our method by two simulations of food and fashion applications because there are services focusing on these domains. 

For food recognition, we used FOOD-101 \cite{bossard14}, which is a real-world food dataset containing the 101 most popular and consistently named dishes collected from \url{foodspotting.com}. FOOD-101 consists of 750 images per class for training and 250 images per class for testing. The training images of FOOD-101 \cite{bossard14} are not cleaned and contain some amount of noise. We evaluated our method on FOOD-101 \cite{bossard14} as ID and TinyImageNet-crop (TINc)/LSUN-crop (LSUNc) as OOD.

For fashion recognition, we used DeepFashion \cite{liuLQWTcvpr16DeepFashion}, a large-scale clothes dataset. We used the Category and Attribute Prediction Benchmark dataset of DeepFashion \cite{liuLQWTcvpr16DeepFashion}, which consists of 289,222 images of clothes and 50 clothing classes. We used DeepFashion \cite{liuLQWTcvpr16DeepFashion} as ID and TinyImageNet-crop (TINc)/LSUN-crop (LSUNc) as OOD.

We resized the FOOD-101 \cite{bossard14} and DeepFashion \cite{liuLQWTcvpr16DeepFashion} images to $32 \times 32$. For FOOD-101 \cite{bossard14}, the original train split was used as $X_{in}$; 1,000 images from the original test split were used for validation and the remaining test images were used as $X_{ul}$. For DeepFashion \cite{liuLQWTcvpr16DeepFashion}, the original train split was used as $X_{in}$; 1,000 images from the original validation split were used for validation and the original test images were used as $X_{ul}$ for unsupervised training and evaluation.     
\vspace{-7pt}
\subsubsection{Experimental Results}
\vspace{-7pt}
\Tref{tbl:real_result} shows the comparison of our method, ODIN \cite{liang2017enhancing} and ELOC \cite{vyas132018out} on real-world simulation datasets. These results clearly show that our architecture significantly outperforms the other existing methods, ODIN \cite{liang2017enhancing} and ELOC \cite{vyas132018out}, by a considerable margin on all datasets. Furthermore, our method nearly perfectly detects non-food and non-fashion images.
\vspace{-5pt}
\section{Conclusion}
\vspace{-5pt}
In this paper, we proposed a novel approach for detecting out-of-distribution data samples in neural networks, which utilizes two classifiers to detect OOD samples that are far from the support of the ID samples. Our method does not require labeled OOD samples to train the neural network. We extensively evaluated our method not only on OOD detection benchmarks, but also on real-world simulation datasets. Our method significantly outperformed the current state-of-the-art methods on different DNN architectures across various in and out-of-distribution dataset pairs.
\vspace{-15pt}
\section{Acknowledgements} 
\vspace{-5pt}
This work was partially supported by JST CREST JPMJCR1686 and JSPS KAKENHI 18H03254, Japan.

{\small
\bibliographystyle{ieee_fullname}
\bibliography{egbib}
}

\end{document}